\documentclass[twoside]{article}
\usepackage[accepted]{aistats2019}

\usepackage[utf8]{inputenc} 
\usepackage[T1]{fontenc}    
\usepackage{hyperref}       
\usepackage{url}            
\usepackage{booktabs}       
\usepackage{amsfonts}       
\usepackage{nicefrac}       
\usepackage{microtype}      

\usepackage[round]{natbib}

\usepackage{amssymb}
\usepackage{amsmath}
\DeclareMathOperator*{\argmax}{\arg\!\max}

\usepackage{graphicx}
\usepackage[algoruled,boxed]{algorithm2e}
\usepackage{textcomp}
\usepackage{subcaption}
\usepackage{xcolor}
\begin{document}

\runningauthor{Lionel~Blondé, Alexandros~Kalousis}

\twocolumn[
\aistatstitle{Sample-Efficient Imitation Learning
    via Generative Adversarial Nets}
\aistatsauthor{
    Lionel~Blondé \\
    \texttt{lionel.blonde@etu.unige.ch}
    \And
    Alexandros~Kalousis \\
    \texttt{alexandros.kalousis@hesge.ch}
}
\aistatsaddress{
    University of Geneva, Switzerland 
    \And
    Geneva School of Business Administration, HES-SO
}
]

\begin{abstract}
GAIL is a recent successful imitation learning architecture
that exploits the adversarial training procedure introduced in GANs.
Albeit successful at generating behaviours similar to those demonstrated
to the agent, GAIL suffers from a high sample complexity in the number of
interactions it has to carry out in the environment in order to achieve
satisfactory performance.
We dramatically shrink the amount of interactions with the
environment necessary to learn well-behaved imitation policies,
by up to several orders of magnitude.
Our framework, operating in the model-free regime,
exhibits a significant increase in sample-efficiency over previous methods
by simultaneously
a) learning a self-tuned adversarially-trained surrogate reward and
b) leveraging an off-policy actor-critic architecture.
We show that our approach is simple to implement
and that the learned agents remain remarkably stable,
as shown in our experiments that span a variety of continuous control tasks.
Video visualisations available at:
\url{https://youtu.be/-nCsqUJnRKU}.
\end{abstract}

\section{Introduction}

Reinforcement learning (RL) is a powerful and extensive framework enabling a
learner to tackle complex continuous control tasks \citep{Sutton1998-ow}.
Leveraging strong function approximators such as multi-layer neural networks,
deep reinforcement learning alleviates the customary preliminary workload
consisting in hand-crafting relevant features for the learning agent to work
on.
While being freed from this engineering burden opens up the framework to an
even broader range of complex control and planning tasks, RL remains hindered
by its reliance on reward design, referred to as
\textit{reward shaping}.
Albeit appealing in theory, shaping often requires an intimidating amount of
engineering via trial and error to yield natural-looking behaviours and makes
the system prone to premature convergence to local minima \citep{Ng1999-lv}.

Imitation learning breaks free from the preliminary reward function
hand-crafting step as it does not need access to a reinforcement signal.
Instead, imitation learning learns to perform a task directly from expert
demonstrations.
The emerging policies mimic the behaviour displayed by the expert in those
demonstrations.
Learning from demonstrations (LfD) has enabled significant advances in robotics
\citep{Billard2008-jb} and autonomous driving \citep{Pomerleau1989-nh,
Pomerleau1990-lm}.
Such models were fit from the expert demonstrations alone in a supervised
fashion, without gathering new data in simulation.
Albeit efficient when data is abundant, they tend to be frail as the agent
strays from the expert trajectories.
The ensuing compounding of errors causes a covariate shift \citep{Ross2010-eb,
Ross2011-dn}.
This approach, referred to as \textit{behavioral cloning}, is therefore poorly
adapted
for imitation.
Those limitations stem from the sequential nature of the problem.

The caveats of behavioral cloning have recently been successfully addressed by
Ho and Ermon \citep{Ho2016-bv} who introduced a model-free imitation learning
method called \textit{Generative Adversarial Imitation Learning} (GAIL).
Leveraging Generative Adversarial Networks (GAN) \citep{Goodfellow2014-yk},
GAIL alleviates the limitations of the supervised approach by
a) learning a reward surrogate that explains the behaviour shown in the
demonstrations and
b) following an RL procedure in an inner loop, consisting in performing
rollouts in a simulated environment with the learned surrogate as
reinforcement signal.
Several works have built on GAIL to overcome the weaknesses it inherits from
GANs, with a particular emphasis on avoiding mode collapse
\citep{Li2017-sb, Hausman2017-hb, Kuefler2017-zu},
causing policies to fail at displaying the diversity of demonstrated
behaviours or skills \citep{Goodfellow2017-pv}.
However, as the authors point out
in the original paper (\citep{Ho2016-bv}, \textsc{Section} 7),
GAIL suffers from severe sample inefficiency.
It is this limitation of GAIL that we address in this paper.
``Sample-efficient'' here means that we focus on limiting the number of
agent-environment interactions, in contrast with reducing the number of
demonstrations needed by the agent.
Although learning from fewer demonstrations is not the primary focus of this work,
our experiments span a spectrum of demonstration dataset sizes.

Failures of previous works to address the exceeding sample complexity stems
from the on-policy nature of the RL procedure they employ.
In such methods, every interaction in a given rollout typically
is used to compute the Monte Carlo estimate of the state value
by summing the rewards accumulated during the current trajectory.
The experienced transitions are then discarded.
Holding on to past trajectories to carry out more than a single optimization
step might appear viable but often results
in destructively large policy updates \citep{Schulman2017-ou}.
Gradients based on those estimates therefore suffer from high variance, which
can be reduced by sampling more intensively, hence the deterring sample
complexity.

In this work, we introduce a novel method that successfully addresses the
impeding sample inefficiency in the number of simulator queries
suffered by previous methods.
By designing an off-policy learning procedure relying on the use of
retained past experiences, we considerably shrink the amount of
interactions necessary to learn good imitation policies.
Despite involving an adversarial training procedure and an actor-critic method,
both notorious
for being prone to instabilities and prohibitively difficult to train,
our technique demonstrates consistent stability,
as shown in the experimental section.
Additionally, our reliance on the deterministic policy gradients
allows us to exploit further information about the learned reward
function, such as its gradient.
Previous methods either ignore it by treating the reward signal as a scalar in
a model-free fashion or train a forward model to exploit it.
Our method achieves the best of both worlds as it can perform a backward pass
from the discriminator to the generator (policy) while remaining model-free.

\section{Related Work}

Imitation learning aims to learn how to perform tasks solely from expert
demonstrations.
Two approaches are typically adopted to tackle imitation learning problems:
a) \textit{behavioral cloning} (BC)
\citep{Pomerleau1989-nh, Pomerleau1990-lm}, which
learns a policy via regression on the state-action pairs from the expert
trajectories, and
b) \textit{apprenticeship learning} (AL)
\citep{Abbeel2004-rb}, which posits the existence of some unknown reward
function under which the expert policy is optimal and learns a policy by
i) recovering the reward that the expert is assumed to maximise
(an approach called \textit{inverse reinforcement learning} (IRL)) and
ii) running an RL procedure with this recovered signal.
As a supervised approach, BC is limited to the available demonstrations to
learn a regression model, whose predictions worsen dramatically as the agent
strays from the demonstrated trajectories.
It then becomes increasingly difficult for the model to recover as the errors
compound \citep{Ross2010-eb, Ross2011-dn, Bagnell2015-ni}.
Only the presence of correcting behaviour in the demonstration dataset can
allow BC to produce robust policies.
AL alleviates this weakness by entangling learning the reward function and
learning the mimicking policy, leveraging the return of the latter to adjust
the parameters of the former. Models are trained on traces of interaction with
the environment rather than on a fixed state pool, leading to greater
generalization to states absent from the demonstrations.
Albeit preventing errors from compounding, IRL comes with a high computational
cost, as both modelling the reward function and solving the ensuing RL problem
(per learning iteration) can be resource intensive
\citep{Syed2008-zo, Syed2008-su, Ho2016-xn, Levine2011-hi}.

In an attempt to overcome the shortcomings of IRL, Ho and Ermon
\citep{Ho2016-bv} managed to bypass the need for learning the reward function
assumed to have been optimised by the expert when collecting the
demonstrations.
The proposed approach to AL,
\textit{Generative Adversarial Imitation Learning} (GAIL),
relies on an essential step consisting in learning
a surrogate function measuring the similarity between the learned policy and
the expert policy, using Generative Adversarial Networks (GAN)
\citep{Goodfellow2014-yk}.
The learned similarity metric is then employed as a reward proxy to carry out
the RL step, inherent to the AL scheme.
Recently, connections have been drawn between GANs, RL \citep{Pfau2016-ft} and
IRL \citep{Finn2016-uj}.
In this work, we extend GAIL to further exploit the connections between those
frameworks and overcome a limitation that was left unaddressed: the
burdensome sample inefficiency of the method.

GANs
involve a generator and a discriminator, each represented by a neural network,
making the associated computational graph fully differentiable.
The gradient of the discriminator with respect to the output of
the generator is of primary importance as it indicates how the
generator should change its output to have better chances at fooling the
discriminator at the next iteration.
In GAIL, the generator's role is carried out by a stochastic policy, causing
the computational graph to no longer be differentiable end-to-end.
Following a model-based approach, \citep{Baram2017-es} recovers the
gradient of the discriminator with respect to actions
(via reparametrization tricks)
and with respect to states (via a forward model),
making the computational graph fully differentiable.
In contrast, the method introduced in this work can,
by operating over deterministic policies and leveraging
the deterministic policy gradient theorem \citep{Silver2014-dk},
directly wield the gradient of the discriminator
with respect to the actions,
without requiring gradient estimation techniques
(e.g.~reparametrization trick \citep{Kingma2013-hf},
Gumbel-Softmax trick \citep{Jang2017-tu, Maddison2017-gk}).
Since we stick to the model-free setting,
states remain stochastic nodes and therefore block (backward) gradient flows.

An independent endeavour to overcome the data inefficiency of GAIL has
very recently
been reported in \citep{Kostrikov2018-jo},
in which the authors leverage a similar architecture,
yet rely on an arguably ad-hoc preliminary preprocessing technique on
the demonstrations before the imitation begins.
In contrast, our method does not rely on any preprocessing to
yield gains in sample efficiency by orders of magnitude.

\section{Background}

\paragraph{Setting}

We address the problem of an agent learning to act in an environment in order
to reproduce the behaviour of an expert demonstrator.
No direct supervision is provided to the agent --- she is never directly told
what the optimal action is --- nor does she receives a reinforcement signal
from the environment upon interaction.
Instead, the agent is provided with a pool of trajectories and must use them to
guide its learning process.

\paragraph{Preliminaries}

We model this sequential interactive problem over discrete timesteps as a
\textit{Markov decision process} (MDP) $\mathbb{M}$, formalised as a tuple
$(\mathcal{S}, \mathcal{A}, \rho_0, p, r, \gamma)$.
$\mathcal{S}$ and $\mathcal{A}$ respectively denote the state and action
spaces.
The dynamics are defined by a transition distribution with conditional
density $p(s_{t+1} | s_t, a_t)$, along with $\rho_0$, the density
of the distribution from which the initial state is sampled.
Finally, $\gamma \in (0, 1]$ denotes the discount factor and
$r: \mathcal{S} \times \mathcal{A} \rightarrow \mathbb{R}$ the reward function.
We consider only the fully-observable case, in which the current state can be
described with the current observation $o_t = s_t$, alleviating the need to
involve the entire history of observations.
Although our results are presented following the previous infinite-horizon MDP,
the MDPs involved in our experiments are \textit{episodic},
with $\gamma = 0$ at episode termination.
In the theory, whenever we omit the discount factor, we implicitly assume the
existence of an absorbing state along any agent-generated trajectory.

We formalise the sequential decision making process of the agent by defining a
parameterised policy $\pi_\theta$, modelled via a neural network with parameter
$\theta$.
$\pi_\theta(a_t|s_t)$ designates the conditional probability density
concentrated at action $a_t$ when the agent is in state $s_t$.
In line with our setting, the agent interacts with $\mathbb{M}^-$,
an MDP comprising every element of $\mathbb{M}$ except its reward function $r$.
Since our approach involves learning a surrogate reward function, we use
$\mathbb{M}^+$ to denote the MDP resulting from the augmentation of
$\mathbb{M}^-$ with the learned reward.
We can therefore equivalently assume that the agent interacts with
$\mathbb{M}^+$.
\textit{Trajectories} are traces of interaction between an agent and an MDP.
Specifically, we model trajectories as sequences of \textit{transitions}
$(s_t, a_t, r_t, s_{t+1})$, atomic units of interaction.
\textit{Demonstrations} are provided to the agent through a set of expert
trajectories $\tau_e$, generated by an expert policy $\pi_e$ in $\mathbb{M}$.

We now introduce
additional concepts and notations that will be used in the
remainder of this work.
The \textit{return} is the total discounted reward from timestep $t$ onwards:
$R_t^\gamma \triangleq \sum_{k=t}^{+\infty} \gamma^{k-t} r(s_k, a_k)$.
The state-action value, or Q-value, is the expected return after
picking action $a_t$ in state $s_t$, and thereafter following policy
$\pi_\theta$:
$Q^{\pi_\theta}(s_t, a_t) \triangleq
\mathbb{E}_{\pi_\theta}^{>t}[R_t^\gamma]$,
where $\mathbb{E}_{\pi_\theta}^{>t}[\cdot]$ denotes the expectation taken along
trajectories generated by $\pi_\theta$ in $\mathbb{M}^+$ (respectively
$\mathbb{E}_{\pi_e}^{>t}[\cdot]$ for $\pi_e$ in $\mathbb{M}$) and looking
onwards from state $s_t$ and action $a_t$.
We want our agent to find a policy $\pi_\theta$ that maximises the expected
return from the start state, which constitutes our performance objective,
$J(\pi) \triangleq \mathbb{E}_\pi[R_0^\gamma]$,
i.e.~$\pi_\theta = \argmax_\pi \, J(\pi)$.
To ease further notations, we finally introduce the
\textit{discounted state visitation distribution}
of a policy $\pi$, denoted by $\rho^\pi: \mathcal{S} \to [0,1]$, and defined by
$\rho^\pi(s) \triangleq
\sum_{t=0}^{+\infty} \gamma^t \mathbb{P}_{\rho_0, \pi}[s_t = s]$,
where $\mathbb{P}_{\rho_0, \pi}[s_t = s]$
is the probability of arriving at state $s$ at time step $t$ when
sampling the initial state from $\rho_0$ and thereafter following policy $\pi$.
In our experiments, we omit the discount factor  for state visitation, in line
with common practices.

\paragraph{\textsc{Gail}}
Leveraging \textit{Generative Adversarial Networks} \citep{Goodfellow2014-yk},
\textit{Generative Adversarial Imitation Learning} \citep{Ho2016-bv}
introduces an extra neural network $D_\phi$
to play the role of \textit{discriminator}, while the role of
\textit{generator} is carried out by the agent's policy $\pi_\theta$.
$D_\phi$ tries to assert whether a given state-action pair
originates from trajectories of $\pi_\theta$ or $\pi_e$, while $\pi_\theta$
attempts to fool $D_\phi$ into believing her state-action pairs come from
$\pi_e$.
The situation can be described as a minimax problem
$\min_\theta \max_\phi \, V(\theta, \phi)$, where
the \textit{value} of the two-player game is
$V(\theta, \phi) \triangleq
\mathbb{E}_{\pi_\theta}[\log (1 - D_\phi(s, a))]
+ \mathbb{E}_{\pi_e}[\log D_\phi(s, a)]$.
We omit the causal entropy term for brevity.
The optimization is however hindered by the stochasticity
of $\pi_\theta$,
causing $V(\theta, \phi)$ to be non-differentiable with respect to $\theta$.

The solution proposed in \citep{Ho2016-bv} consists in alternating between a
gradient step (\textsc{Adam}, \citep{Kingma2014-op}) on $\phi$ to increase
$V(\theta, \phi)$ with respect to $D_\phi$, and a policy optimization step
(TRPO, \citep{Schulman2015-jt}) on $\theta$ to decrease $V(\theta, \phi)$ with
respect to $\pi_\theta$.
In other words, while $D_\phi$ is trained as a binary classifier to predict if
a given state-action pair is real (from $\pi_e$) or generated (from
$\pi_\theta$), the policy $\pi_\theta$ is trained by being rewarded for
successfully confusing $D_\phi$ into believing that generated samples are coming
from $\pi_e$, and treating this reward as if it were an external
analytically-unknown reward from the environment.

\paragraph{Actor-critic}
Policy gradient methods with function approximation \citep{Sutton1999-ii},
referred to as \textit{actor-critic} (AC) methods,
interleave \textit{policy evaluation} with \textit{policy iteration}.
Policy evaluation estimates the state-action value function with a function
approximator called \textit{critic}
$Q_\psi \approx Q^{\pi_\theta}$, usually via either
Monte-Carlo (MC) estimation or Temporal Difference (TD) learning.
Policy iteration updates the policy $\pi_\theta$ by greedily optimising it
against the estimated critic $Q_\psi$.

\section{Algorithm}

The approach in this paper,
named \textit{Sample-efficient Adversarial Mimic} (\textsc{Sam}),
adopts an off-policy TD learning paradigm.
By storing past experiences and replaying them in an uncorrelated fashion,
\textsc{Sam} displays significant gains in sample-efficiency,
in line with \citep{Wang2016-mp, Gu2016-uc}.
To solve the differentiability bottleneck of \citep{Ho2016-bv}
caused by the stochasticity of its generator,
we operate over deterministic policies.
At a given state $s_t$, following its deterministic policy $\mu_\theta$,
an agent selects the action $a_t = \mu_\theta(s_t)$.
Alternatively, we can obtain a deterministic policy from any stochastic policy
$\pi_\theta$
by systematically picking the average action for a given state:
$\mu_\theta(s_t) = \mathbb{E}_a[\pi_\theta(a | s_t)]$.
By relying on an off-policy actor-critic architecture and
wielding deterministic policies,
\textsc{Sam} builds on
the \textit{Deep Deterministic Policy Gradients} (DDPG) algorithm
\citep{Lillicrap2016-xa},
in the context of Imitation Learning.

\begin{figure}[!h]
  \center\scalebox{0.28}[0.28]{\includegraphics{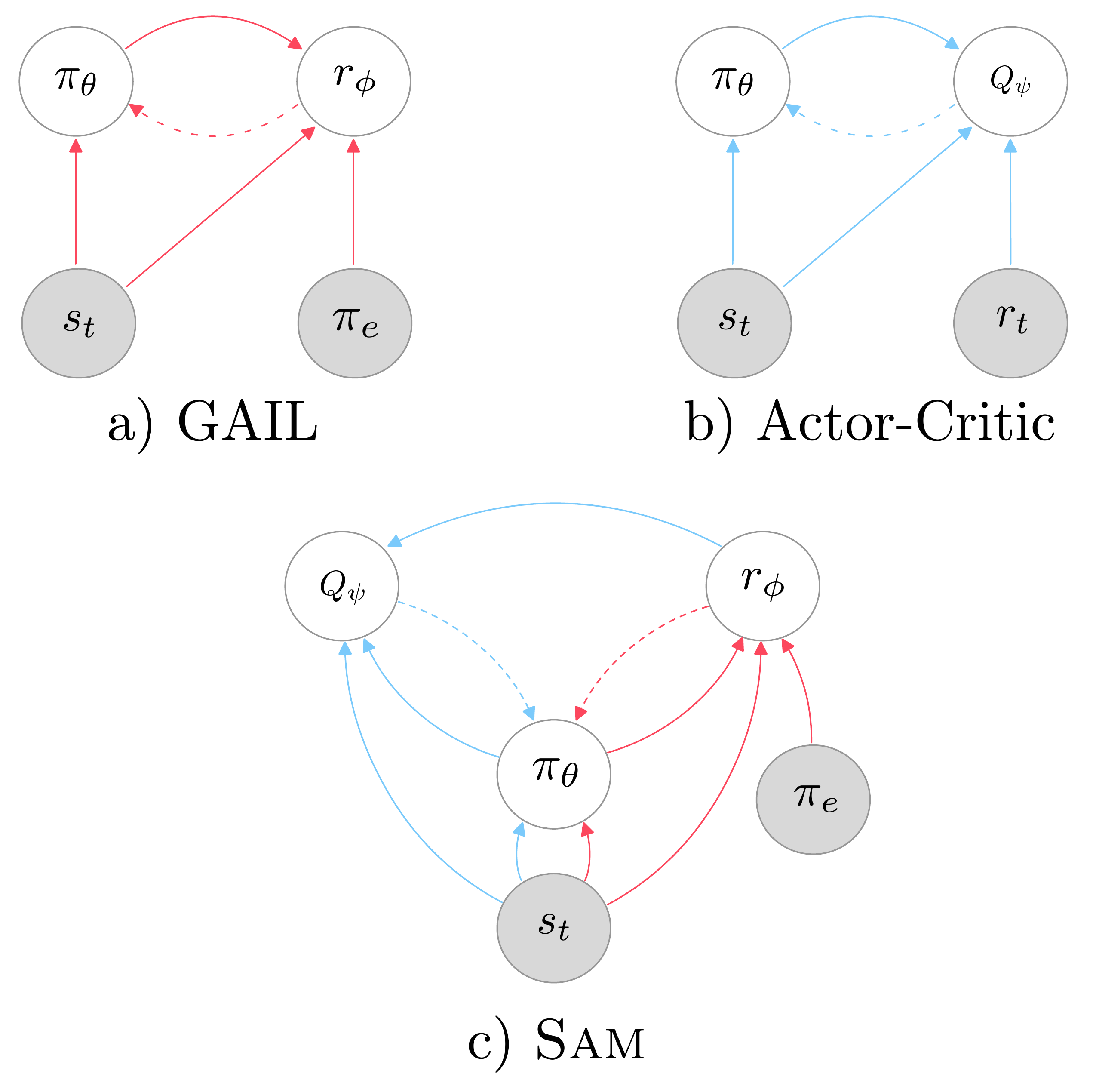}}
  \caption{
   Inter-module relationships in different neural architectures
    (the scope of this figure was inspired from \citep{Pfau2016-ft}).
    Modules with distinct loss functions are depicted with empty circles, while
    filled circles designate environmental entities.
    Solid and dotted arrows respectively represent
    (forward) flow of information
    and
    (backward) flow of gradient.
    a)
    Generative Adversarial Imitation Learning \citep{Ho2016-bv}
    b)
    Actor-Critic architecture \citep{Sutton1999-ii}
    c)
    \textsc{Sam} (this work).
    Note that in \textsc{Sam}, the critic takes in information from the reward
    module, while in the vanilla AC architecture, the critic receives the
    reward from the environment.
    The gradient flow from the critic to the reward module must however
    be sealed.
    Indeed, such
    a gradient flow would allow the policy to adjust its parameters to induce
    values of the reward which yield low TD residuals, hence preventing both
    critic and reward modules to be learned as intended.
  }
  \label{fig:diags}
\end{figure}

\textsc{Sam} is composed of three interconnected learning modules:
a \textit{reward module} (parameter $\phi$),
a \textit{policy module} (parameter $\theta$), and
a \textit{critic module} (parameter $\psi$)
(\textsc{Figure}~\ref{fig:diags}).
The reward and policy modules are both involved in a GAN's adversarial
training procedure,
while the policy and critic modules are trained as an
actor-critic architecture.
As reminded recently in \citep{Pfau2016-ft}, GANs and actor-critic
architectures can be both framed as bilevel optimization problems, each
involving two competing components, which we just listed out for both
architectures.
Interestingly, the policy module plays a role in both problems,
tying the two bilevel optimization problems together.
In one problem, the policy module is trained against the reward module,
while in the other,
the policy module is trained against the critic module.
The reward and critic modules can therefore be seen as serving analogous roles
in their respective bilevel optimization problems:
forging and maintaining a signal which enables the reward-seeking policy to
adopt the desired behaviour.
How each of these component is optimised is described in the
subsequent dedicated sections.

As an off-policy method, \textsc{Sam} cycles through the following steps:
i) the agent uses $\pi_\theta$ to interact with $\mathbb{M}^+$,
ii) stores the experienced transitions $\mathcal{C}$ in a replay buffer $\mathcal{R}$,
iii) updates the reward module $\phi$ with an equal mixture of uniformly sampled
state-action pairs
from $\mathcal{C}$ and $\tau_e$,
iv) updates the reward module $\phi$ with an equal mixture of uniformly sampled
state-action pairs
from $\mathcal{R}$ and $\tau_e$, and
v) updates the policy module $\theta$ and critic module $\psi$
with transitions sampled from $\mathcal{R}$.
Note that while sampling uniformly from $\mathcal{C}$ (iii) gives
states and actions distributed as $\rho^{\pi_\theta}$ and $\pi_\theta$ respectively
(on-policy),
sampling uniformly from $\mathcal{R}$ (iv) gives
states and actions distributed as $\rho^\beta$ and $\beta$ respectively,
where $\beta$ denotes the off-policy sampling mixture distribution
corresponding to sampling transitions uniformly from the replay buffer.
A more detailed description of the training procedure is laid out in the
algorithm pseudo-code (\textsc{Algorithm}~\ref{fig:algo}).

\paragraph{Reward}
We introduce a reward network with parameter vector $\phi$,
operating as the discriminator.
The cross-entropy loss used to train the reward network is:
\begin{align}
  \mathbb{E}_{\pi_\theta}[- \log (1 - D_\phi(s, a))]
  & + \mathbb{E}_{\pi_e}[- \log D_\phi(s, a)]
  \\
  & + \lambda \mathfrak{R}_{\text{GP}}(\phi)
  \label{eq:philoss}
\end{align}
where $\mathfrak{R}_{\text{GP}}(\phi)$ is a
penalty on the discriminator gradient,
as introduced in \citep{Gulrajani2017-mr}, \textsc{Section} 4.
\citep{Lucic2017-nz} reports benefits from applying such regulariser to
the non-saturated variant of the discriminator loss, although it
was initially introduced for Wasserstein GANs \citep{Arjovsky2017-la} in
\citep{Gulrajani2017-mr}.
This penalty favours our method by further improving its stability.

The reward is defined as the negative of the generator loss.
The later has been declined in many variants, which are
thoroughly compared in \citep{Lucic2017-nz}.
We can therefore analogously define a synthetic reward for each of these
forms. We go over and discuss major ones in supplementary material.
Additionally, \citep{Fu2018-zu} proposes an extra variant
in the context of IRL.
In the remainder, we use
$r_\phi(s_t, a_t) = - \log (1 - D_\phi(s_t, a_t))$
as synthetic reward.
The reward network is trained, each iteration,
first on the mini-batch most recently collected by $\pi_\theta$,
then on mini-batches sampled from the replay buffer.
Although \citep{Pfau2016-ft} reports that using a replay buffer in GANs
causes the generation to be poor,
we do not seem to suffer the same detrimental effect in
the continuous control tasks we tackle.

\paragraph{Critic}
The loss optimised by the critic, noted $\ell(\psi)$, involves three
components:
i) a $1$-step Bellman residual $\ell_1(\psi)$,
ii) a $n$-step Bellman residual $\ell_n(\psi)$, and
iii) a weight decay regulariser $\mathfrak{R}_{\text{WD}}(\psi)$.
A similar loss is employed in \citep{Vecerik2017-ue} in the context of
Reinforcement Learning from Demonstrations.
While the authors use weight decay regularisers for both the policy and
the critic, we restrain from decaying the policy's weights since,
in our setting, the policy plays a role in two distinct optimization problems.
We do not apply a weight decay regulariser for the discriminator either,
as it was proven to cause the Wasserstein GAN
\textit{critic}
(name given to the discriminator in Wasserstein GANs) to diverge
\citep{Gulrajani2017-mr}.

We define the critic loss as follows:
\begin{align}
  \ell(\psi) =
  \ell_1(\psi) + \ell_n(\psi) + \nu \mathfrak{R}_{\text{WD}}(\psi)
  \label{eq:psiloss}
\end{align}
where $\nu$ is a hyperparameter that determines how much decay is used.
The losses i) and ii) are defined respectively based on the $1$-step and
$n$-step lookahead versions of the Bellman equation,
\begin{align}
  \tilde{Q}_\psi^1(s_t, a_t) & \triangleq
  r_\phi(s_t, a_t)
  \\
  & + \gamma Q_\psi(s_{t+1}, \mu_\theta(s_{t+1}))
  \label{eq:qtilde1}
  \\
  \tilde{Q}_\psi^n(s_t, a_t) & \triangleq
  \sum_{k=0}^{n-1} \gamma^k r_\phi(s_{t+k}, a_{t+k})
  \\
  & + \gamma^n Q_\psi(s_{t+n}, \mu_\theta(s_{t+n}))
  \label{eq:qtilden}
\end{align}
yielding the critic losses:
\begin{align}
  \ell_1(\psi) \triangleq
  \mathbb{E}_{s_t \sim \rho^\beta, a_t \sim \beta}
  [
  (\tilde{Q}_\psi^1 - Q_\psi)^2(s_t, a_t)
  ]
  \\
  \ell_n(\psi) \triangleq
  \mathbb{E}_{s_t \sim \rho^\beta, a_t \sim \beta}
  [
  (\tilde{Q}_\psi^n - Q_\psi)^2(s_t, a_t)
  ]
  \label{eq:philosses}
\end{align}
where $\mathbb{E}_{s_t \sim \rho^\beta, , a_t \sim \beta}[\cdot]$
signifies that
transitions are sampled from the replay buffer $\mathcal{R}$,
using in effect the off-policy distribution $\beta$.
Both $Q_\psi$ and $\tilde{Q}_\psi^\cdot$
(\eqref{eq:qtilde1}, \eqref{eq:qtilden})
depend on $\psi$,
which might cause severe instability.
In order to prevent the critic from diverging, we use separate \textit{target}
networks for both policy and critic ($\smash{\theta'}$, $\smash{\psi'}$)
to calculate
$\smash{\tilde{Q}_\psi^\cdot}$,
which slowly track the learned parameters ($\theta$, $\psi$).
In line with results exhibited in the recent ablation study
(\textsc{Rainbow} \citep{Hessel2017-ns})
assessing the influence of the various add-ons of DQN \citep{Mnih2013-rb, Mnih2015-iy}
on its performance,
we studied the influence of two add-ons that were transposable to \textsc{Sam}:
longer TD backups and replay prioritisation.
$n$-step returns
not only played a significant role in improving the sample complexity,
but also had a positive influence on stability in the training regime.
Prioritized Experience Replay \citep{Schaul2016-oj}
however prevented \textsc{Sam} from consistently learning well-behaved policies.
Being already prone to overfitting in its original setting
\citep{Schaul2016-oj},
we conjecture this phenomenon is amplified in our setting since the TD-errors,
instrumental in the priority assignments,
depend on rewards that are themselves learned.
Uniform experience replay
offers greater resilience against oversampling
transitions that have wrongfully been assigned high synthetic rewards by the
adversarially-trained reward module.

\paragraph{Policy}
We update the policy $\mu_\theta$ so as to maximise the
\textit{performance objective},
defined as the expected return from the start state.
To that end, the policy is updated by taking a gradient ascent step along:
\begin{align}
  \nabla_\theta^{(1)} J(\mu_\theta)
  & \approx \mathbb{E}_{s_t \sim \rho^\beta}
  \left[
  \nabla_\theta
  Q_\psi(s_t, \mu_\theta(s_t))
  \right]
  \\
  & = \mathbb{E}_{s_t \sim \rho^\beta}
  \left[
  \nabla_\theta
  \mu_\theta(s_t) \nabla_a Q_\psi(s_t, a)
  |_{a = \mu_\theta(s_t)}
  \right]
  \label{eq:thetagrad2}
\end{align}
where the partial derivative with respect to the state is
ignored since we consider the model-free setting.
This gradient estimation stems from the policy gradient theorem proved by
\citep{Silver2014-dk},
and points towards regions of the parameter space in which the policy
displays high similarity with the demonstrator.

We model the synthetic reward as a parametrised function
that takes a state and an action as inputs.
As such, we can take the derivative of the reward with respect to $\theta$.
By applying the chain rule, we obtain:
\begin{align}
  \nabla_\theta^{(2)} J(\mu_\theta)
  & \approx \mathbb{E}_{s_t \sim \rho^\beta}
  \left[
  \nabla_\theta
  r_\phi(s_t, \mu_\theta(s_t))
  \right]
  \\
  & = \mathbb{E}_{s_t \sim \rho^\beta}
  \left[
  \nabla_\theta
  \mu_\theta(s_t) \nabla_a r_\phi(s_t, a)
  |_{a = \mu_\theta(s_t)}
  \right]
  \label{eq:thetagrad1}
\end{align}
which constitutes another estimate of how to update
the policy parameters $\theta$ to increase the similarity between the
policy and the expert (\citep{Sasaki2018-bz} employs a similar estimate).
Each estimate of how well the agent is behaving,
$r_\phi$ and $Q_\psi$, is trained via a
different policy evaluation method, each presenting its own advantages.
The first is updated by adversarial training, providing an accurate estimate of
the immediate similarity with expert trajectories.
The second is trained via TD learning, enabling longer propagation of
rewards along trajectories and effectively tackling the credit assignment
problem.
While our formulation enables us to use either of these gradient estimates,
$\nabla_\theta^{(1)} J(\mu_\theta)$ is more suited to learn control policies in
environments inducing delayed rewards.
As the continuous control tasks we consider in this paper belong to this
category, we use $\nabla_\theta^{(1)} J(\mu_\theta)$ to update the policy module.
While we could use a mixture of $\nabla_\theta^{(1)} J(\mu_\theta)$ and
$\nabla_\theta^{(2)} J(\mu_\theta)$ we found that the latter had a
detrimental effect on the former, as it prevented the policy to reason
across timesteps, resulting in poor reward propagation.

\paragraph{Exploration} Deterministic policies have zero variance in their
predictions for a given
state, translating to no exploratory behaviour.
The exploration problem is therefore treated independently from how the policy
is modelled, by defining a stochastic policy $\pi_\theta$
from the learned deterministic policy $\mu_\theta$.
In this work, we construct $\pi_\theta$ via the combination of
two fundamentally different techniques:
a) by applying an adaptive perturbation to the learned weights $\theta$
(exploration by noise-injection in parameter space
\citep{Plappert2018-rl, Fortunato2017-af}) and
b) by adding temporally-correlated noise sampled from a Ornstein-Uhlenbeck
process $\mathbb{O}\mathbb{U}$
\citep{Lillicrap2016-xa}, well-suited for control tasks involving
inertia (e.g.~simulated robotics and locomotion tasks).
We denote the obtained policy by
$\pi_\theta \triangleq \mu_{\tilde{\theta}} + \mathbb{O}\mathbb{U}$,
where $\tilde{\theta}$ results from applying a) to $\theta$.
When interacting with the environment, \textsc{Sam}
samples from the conditional distribution $\pi_\theta$,
and stores the collected transitions in the replay buffer $\mathcal{R}$.
An interesting result is that the reward is adversarially trained
on samples coming from the parameter-perturbed policy.
Rather than causing severe divergence, it seems that it positively impacts the
adversarial training procedure.
This observation directly echoes noise-injection techniques from the GAN
literature.
The additive noise applied to the output of our policy (which plays the role
of generator in our architecture)
aligns with
\citep{Arjovsky2017-ne}
who add artificial noise to the inputs of the discriminator (although we
do not perturb expert trajectories).
Furthermore, perturbing $\mu_\theta$ in parameter space
draws strong similarities with \citep{Zhao2017-bs}, in which the authors
add Gaussian noise to the layers of the generator.

\begin{algorithm}
  \SetKwComment{Comment}{}{}
  Initialise replay buffer $\mathcal{R}$ \\
  Initialise network parameters ($\phi$, $\theta$, $\psi$) \\
  Initialise target network parameters ($\theta'$, $\psi'$) as respective
  copies of ($\theta$, $\psi$) \\
  \For{$\text{i} \in 1, \ldots, \text{i}_\text{max}$}{
    \Comment{\textcolor{gray}{\# Interact with environment}}
    \Comment{\textcolor{gray}{\# and store collected transitions}}
    \For{$\text{c} \in 1, \ldots, \text{c}_\text{max}$}{
      Interact with environment following $\pi_\theta$
      and
      collect the experienced transitions in $\mathcal{C}$
      augmented with synthetic rewards
      \\
      Store $\mathcal{C}$ in the replay buffer $\mathcal{R}$
    }
    \For{$\text{t} \in 1, \ldots, \text{t}_\text{max}$}{
      \Comment{\textcolor{gray}{\# Update reward module}}
      \For{$\text{d} \in 1, \ldots, \text{d}_\text{max}$}{
        Sample uniformly a minibatch $\mathcal{B}^c$ of
        state-action pairs pairs
        from $\mathcal{C}$ \\
        Sample uniformly a minibatch $\mathcal{B}_e^c$ of
        state-action pairs
        from the
        expert dataset $\tau_e$,
        with $|\mathcal{B}^c| = |\mathcal{B}_e^c|$ \\
        Update synthetic reward parameter $\phi$ with the equal mixture
        $\mathcal{B}^c \cup \mathcal{B}_e^c$ by following the gradient:
        $
        \hat{\mathbb{E}}_{\mathcal{B}^c}
        [\nabla_\phi \log(1 - D_\phi(s, a))]
        +
        \hat{\mathbb{E}}_{\mathcal{B}_e^c}
        [\nabla_\phi \log D_\phi(s, a)]
        +
        \lambda \nabla_\phi \mathfrak{R}_{\text{GP}}(\phi)
        $
        \\
        \vspace{0.2cm}
        Sample uniformly a minibatch $\mathcal{B}^d$ of
        state-action pairs
        from $\mathcal{R}$ \\
        Sample uniformly a minibatch $\mathcal{B}_e^d$ of
        state-action pairs
        from the
        expert dataset $\tau_e$,
        with $|\mathcal{B}^d| = |\mathcal{B}_e^d|$ \\
        Update synthetic reward parameter $\phi$ with the equal mixture
        $\mathcal{B}^d \cup \mathcal{B}_e^d$ by following the gradient:
        $
        \hat{\mathbb{E}}_{\mathcal{B}^d}
        [\nabla_\phi \log(1 - D_\phi(s, a))]
        +
        \hat{\mathbb{E}}_{\mathcal{B}_e^d}
        [\nabla_\phi \log D_\phi(s, a)]
        +
        \lambda \nabla_\phi \mathfrak{R}_{\text{GP}}(\phi)
        $
      }
      \Comment{\textcolor{gray}{\# Update policy and critic modules}}
      \For{$\text{g} \in 1, \ldots, \text{g}_\text{max}$}{
        Sample uniformly a minibatch $\mathcal{B}^g$ of
        transitions
        from $\mathcal{R}$ \\
        Update policy parameter $\theta$
        by following the gradient:
        $
        \hat{\mathbb{E}}_{\mathcal{B}^g}
        [\nabla_\theta J(\mu_\theta)]
        $
        \\
        Update critic parameters $\psi$ by minimizing critic loss:
        $
        \hat{\mathbb{E}}_{\mathcal{B}^g}
        [\ell(\psi)]
        $
        \\
        Update target network parameters ($\theta'$, $\psi'$)
        to slowly track ($\theta$, $\psi$), respectively
      }
    }
  }
  \caption{Sample-efficient Adversarial Mimic}
  \label{fig:algo}
\end{algorithm}

\section{Results}

\begin{figure}
  \centering
  \begin{subfigure}[t]{0.23\textwidth}
    \centering
    \captionsetup{width=.23\linewidth}
    \includegraphics[width=\linewidth]{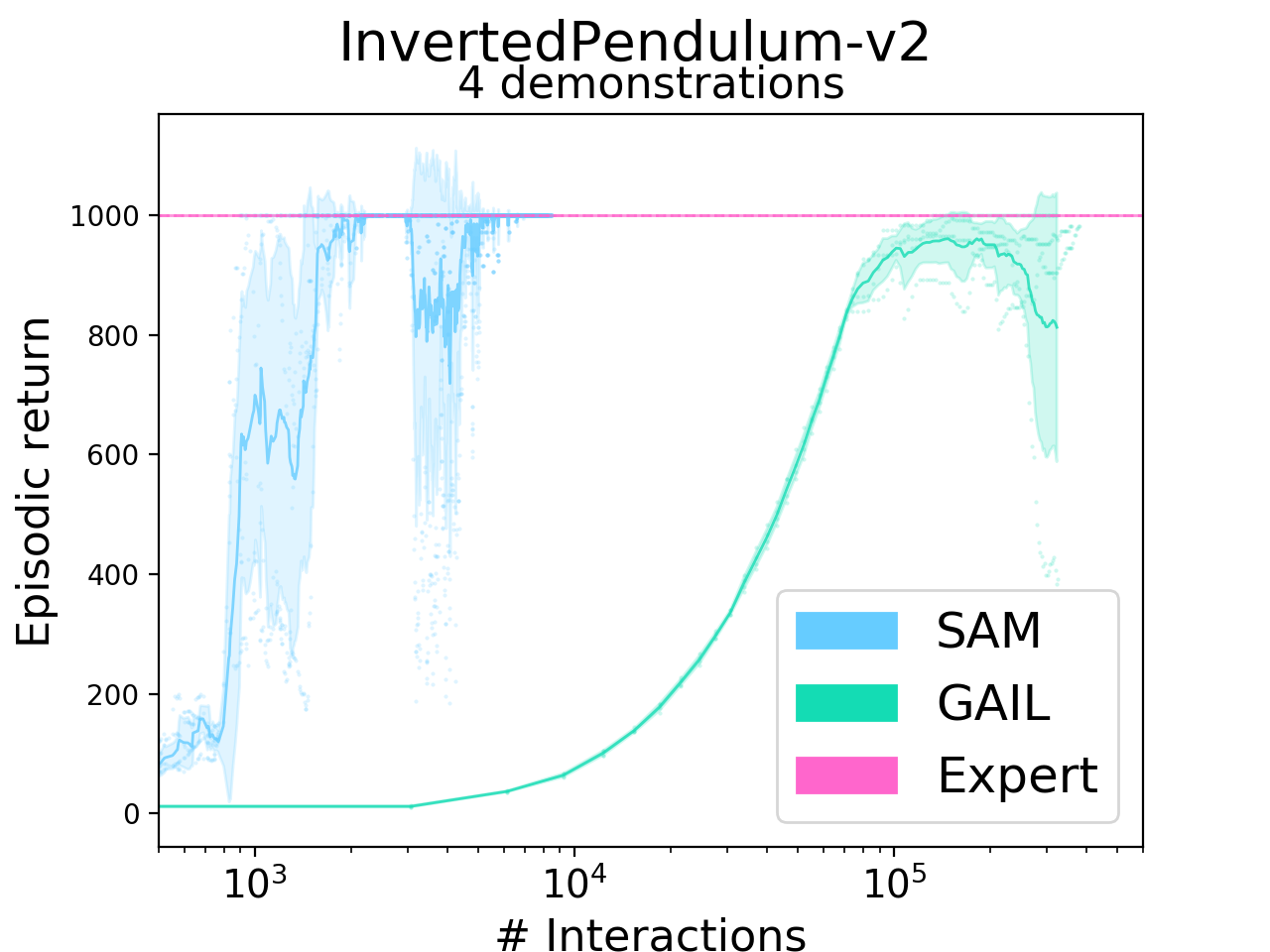}
    \label{}
  \end{subfigure}
  \begin{subfigure}[t]{0.23\textwidth}
    \centering
    \captionsetup{width=.23\linewidth}
    \includegraphics[width=\linewidth]{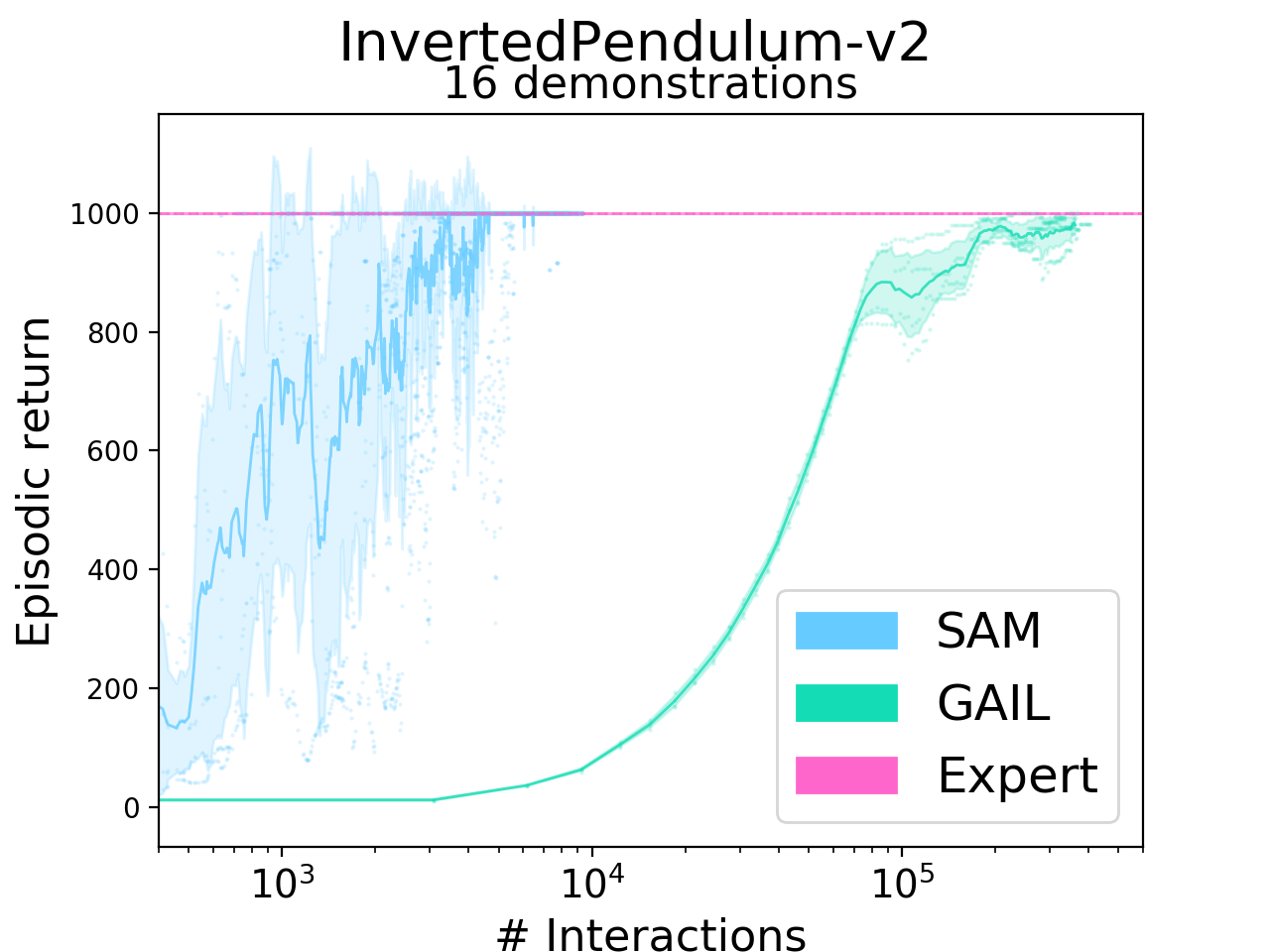}
    \label{}
  \end{subfigure}
  \begin{subfigure}[t]{0.23\textwidth}
    \centering
    \captionsetup{width=.23\linewidth}
    \includegraphics[width=\linewidth]{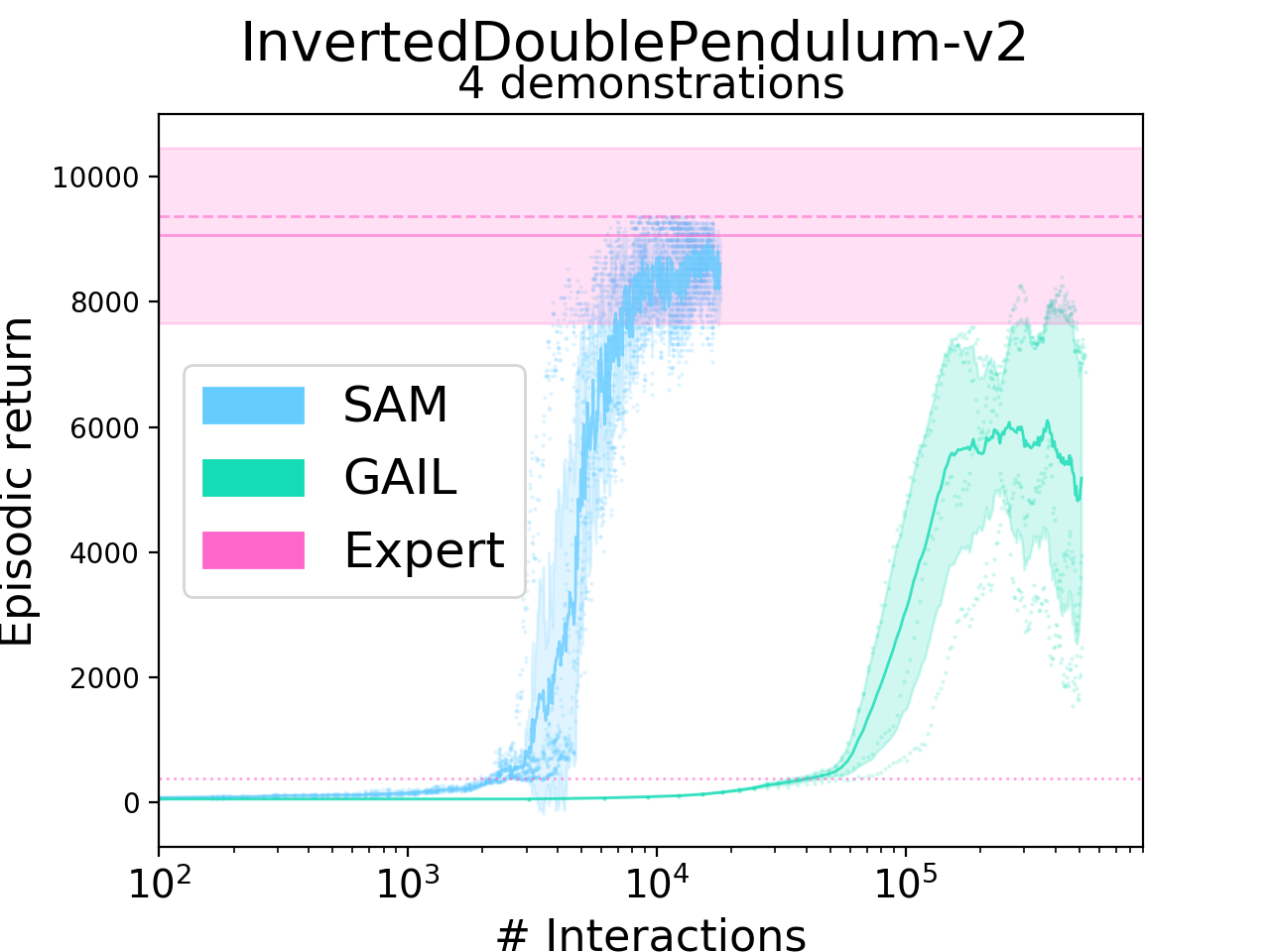}
    \label{}
  \end{subfigure}
  \begin{subfigure}[t]{0.23\textwidth}
    \centering
    \captionsetup{width=.23\linewidth}
    \includegraphics[width=\linewidth]{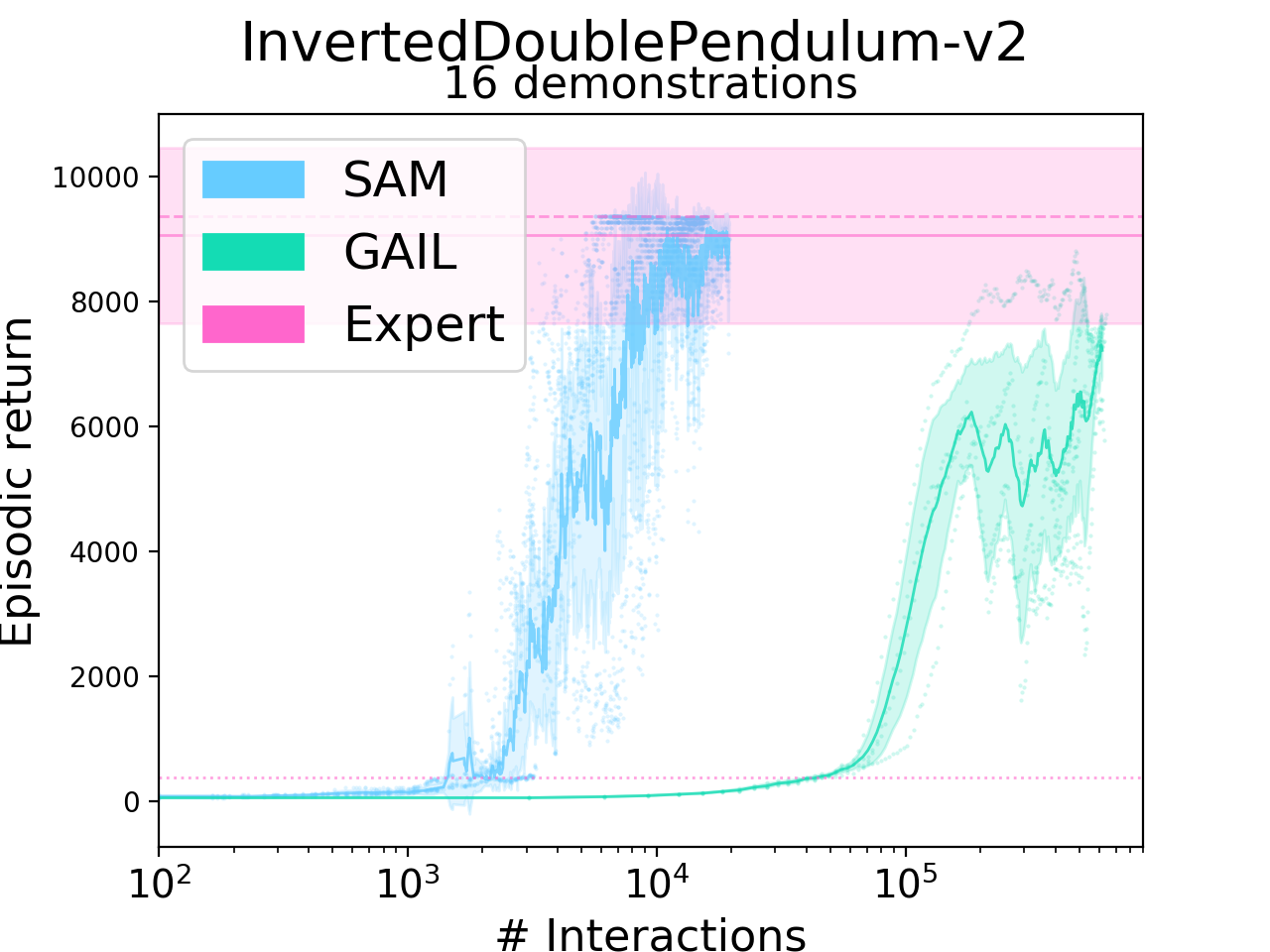}
    \label{}
  \end{subfigure}
  \begin{subfigure}[t]{0.23\textwidth}
    \centering
    \captionsetup{width=.23\linewidth}
    \includegraphics[width=\linewidth]{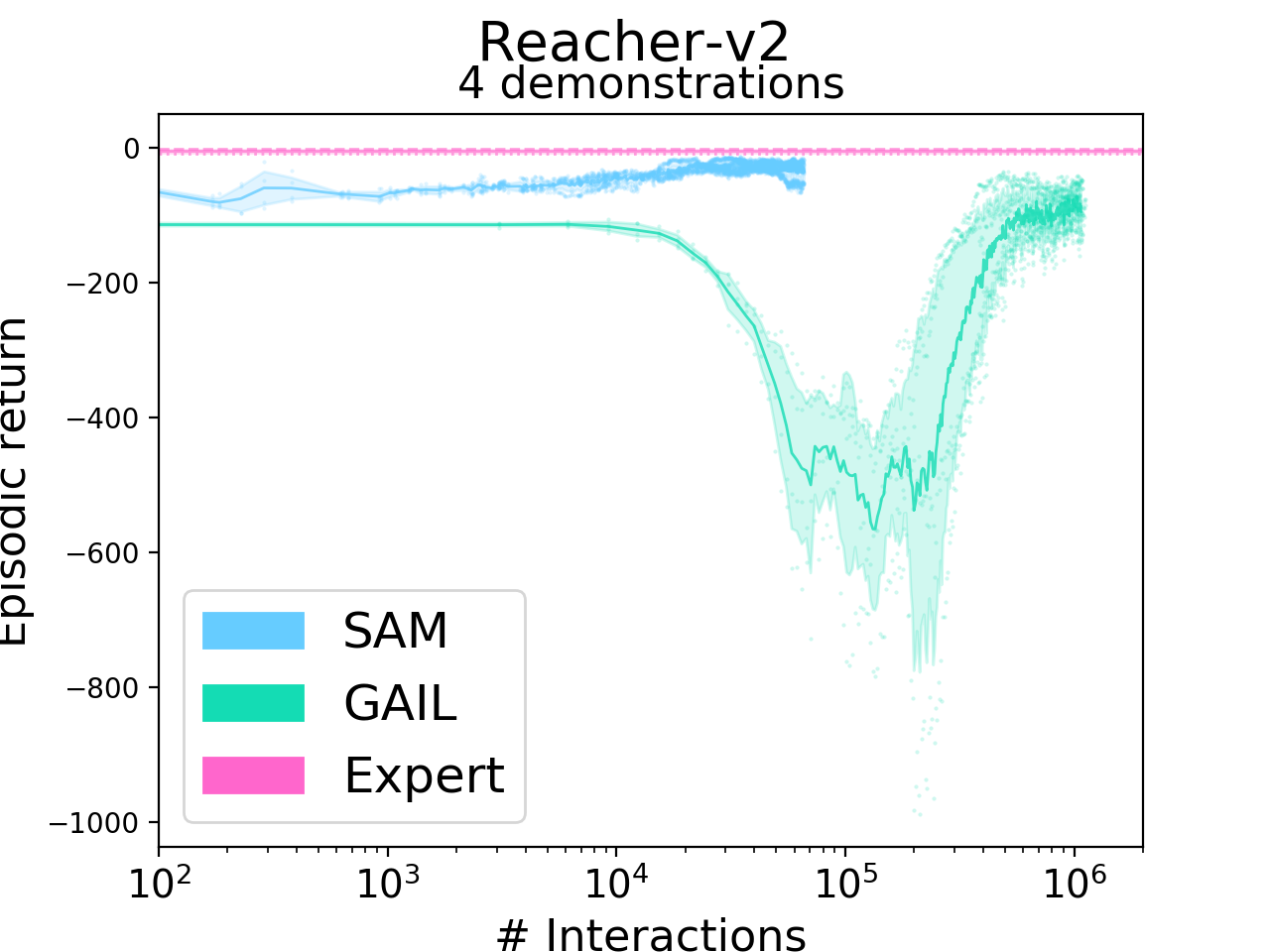}
    \label{}
  \end{subfigure}
  \begin{subfigure}[t]{0.23\textwidth}
    \centering
    \captionsetup{width=.23\linewidth}
    \includegraphics[width=\linewidth]{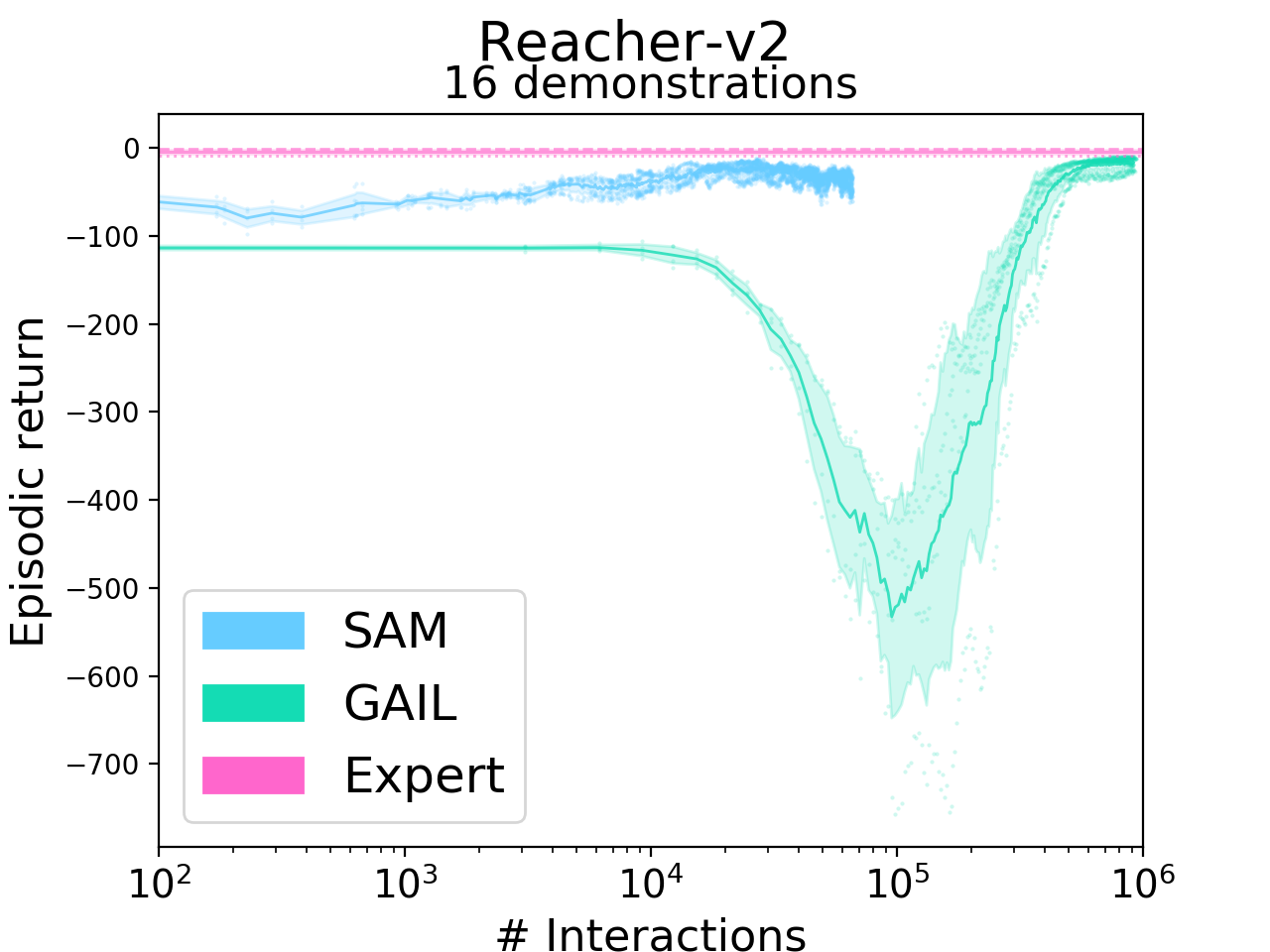}
    \label{}
  \end{subfigure}
  \begin{subfigure}[t]{0.23\textwidth}
    \centering
    \captionsetup{width=.23\linewidth}
    \includegraphics[width=\linewidth]{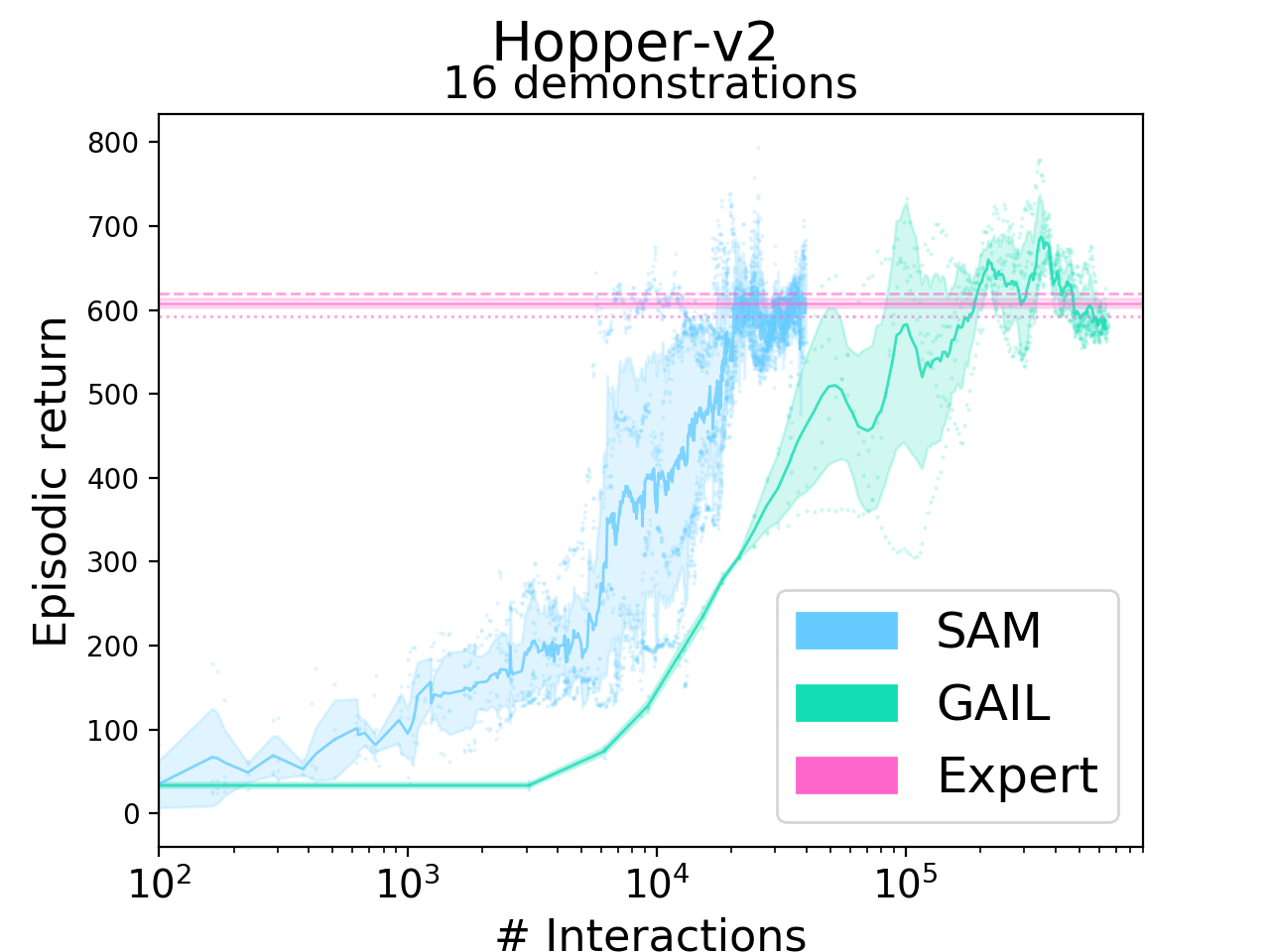}
    \label{}
  \end{subfigure}
  \begin{subfigure}[t]{0.23\textwidth}
    \centering
    \captionsetup{width=.23\linewidth}
    \includegraphics[width=\linewidth]{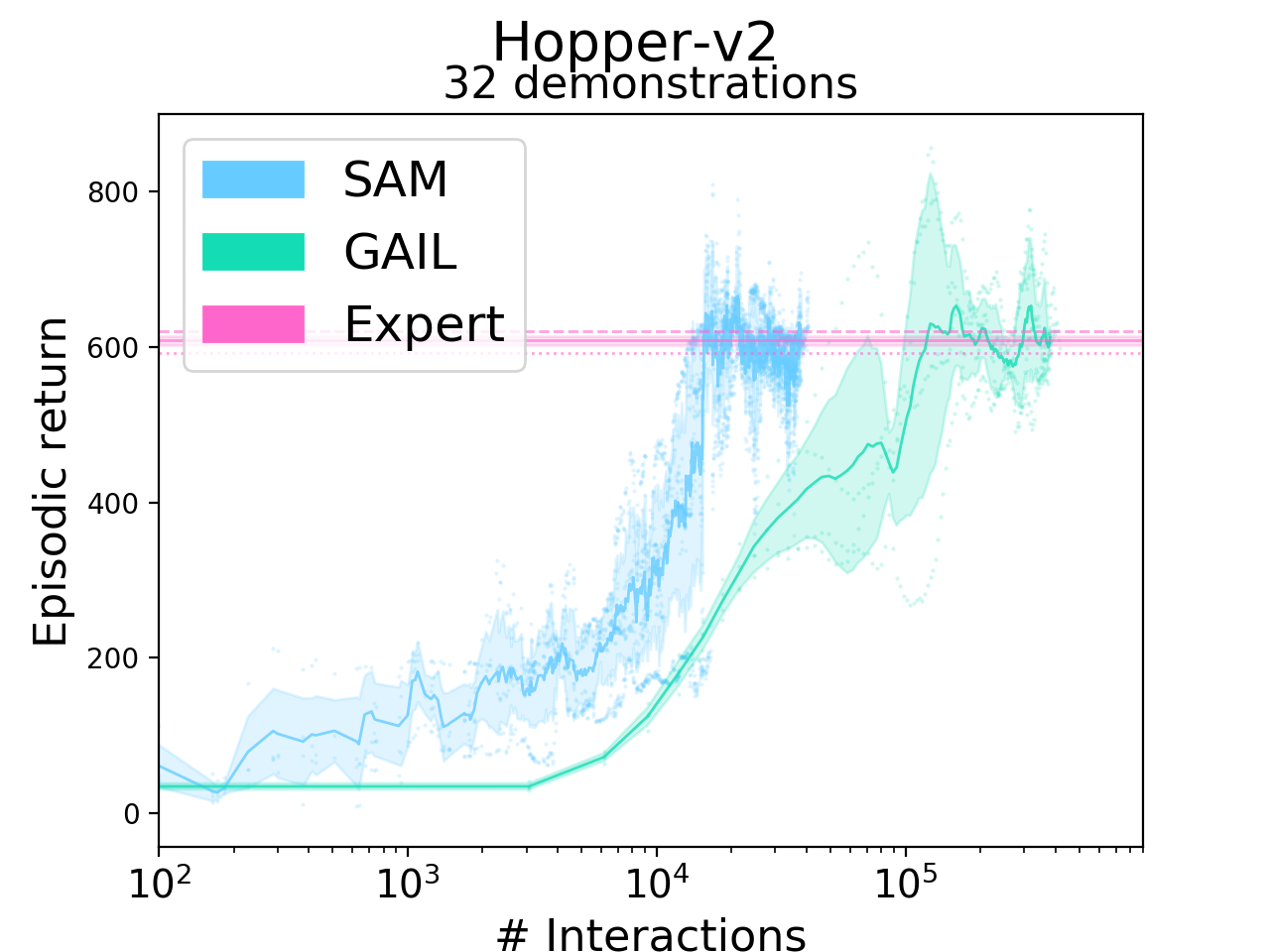}
    \label{}
  \end{subfigure}
  \begin{subfigure}[t]{0.23\textwidth}
    \centering
    \captionsetup{width=.23\linewidth}
    \includegraphics[width=\linewidth]{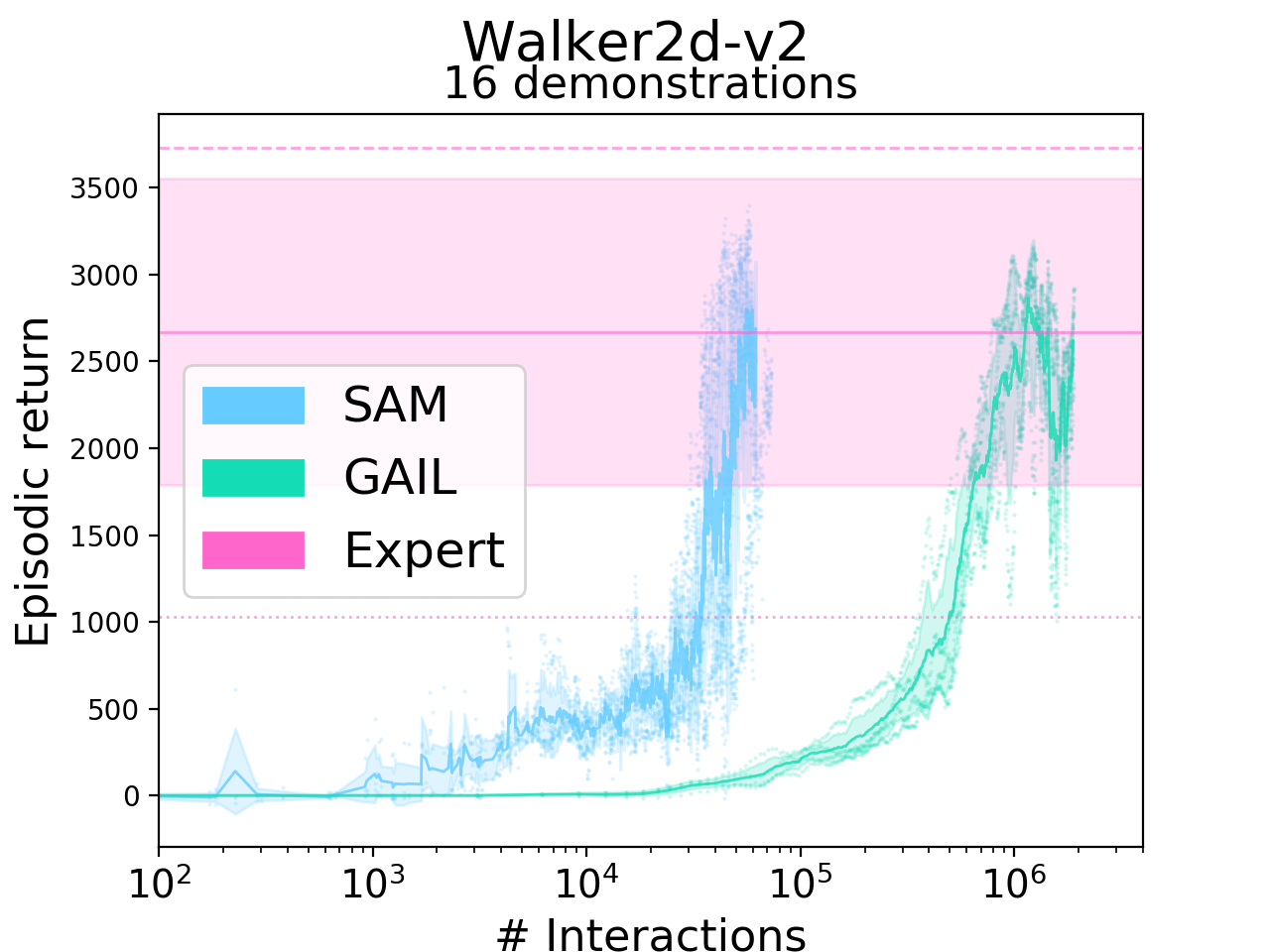}
    \label{}
  \end{subfigure}
  \begin{subfigure}[t]{0.23\textwidth}
    \centering
    \captionsetup{width=.23\linewidth}
    \includegraphics[width=\linewidth]{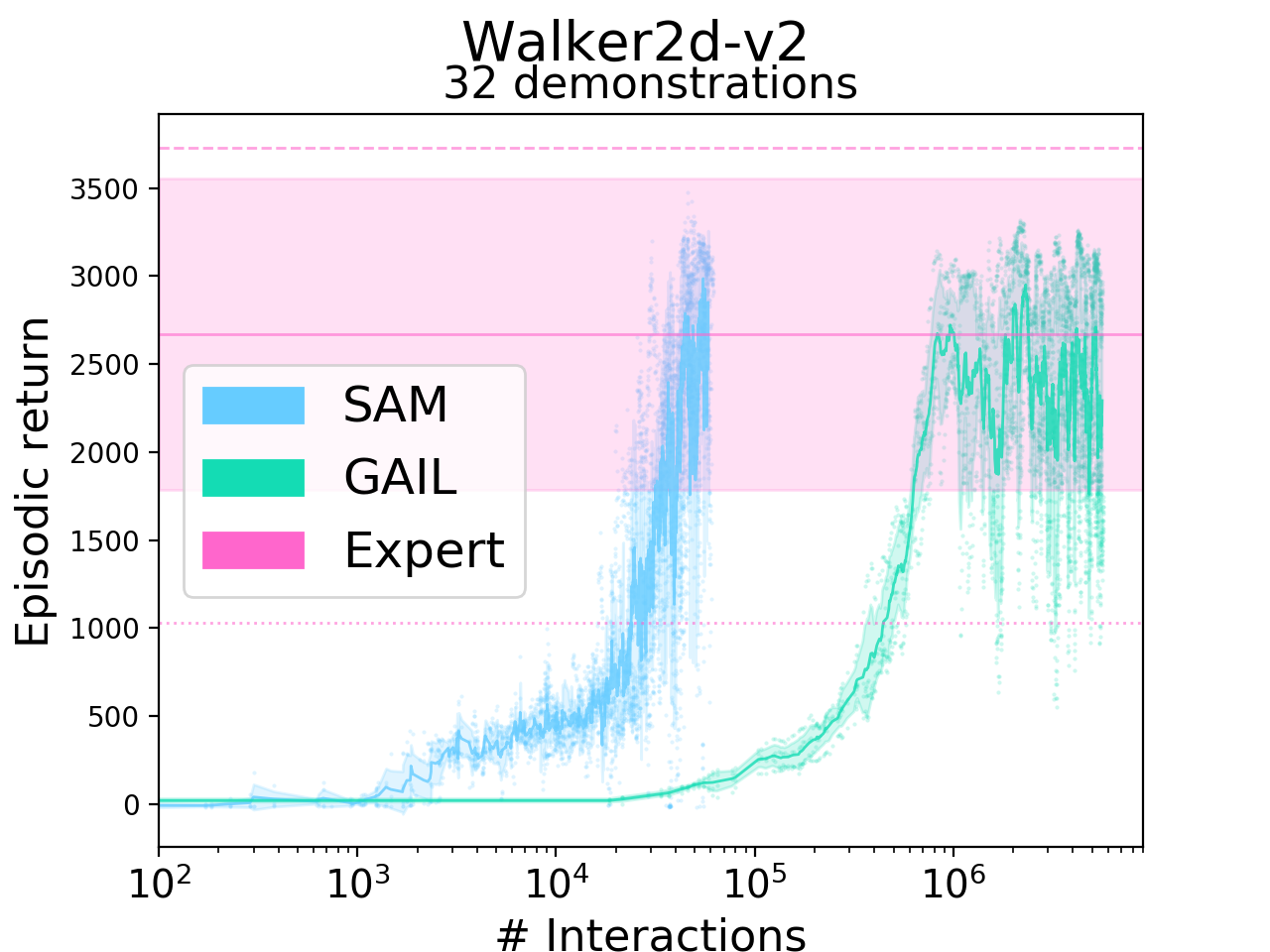}
    \label{}
  \end{subfigure}
  \caption{
    Performance comparison between \textsc{Sam} and GAIL in terms of episodic return.
    The horizontal axis depicts, in logarithmic scale, the number of interactions
    with the environment.
    While there is no ambiguity for GAIL, we used the unperturbed \textsc{Sam}
    policy $\mu_\theta$
    (without parameter noise and additive action noise) to collect those returns
    during a per-iteration evaluation phase.
    The figure shows that our method has a considerably better
    sample-efficiency than GAIL in various continuous control tasks,
    often by several orders of magnitude.
    Red-colored lines and filled areas indicate the performance range of the expert
    demonstrations present in the training set.
    The meaning of the different line styles and colors is given in-text.
  }
  \label{fig:resplots}
\end{figure}

Our agents were trained in physics-based control environments,
built with the \textsc{MuJoCo} physics engine \citep{Todorov2012-gc},
and wrapped via the OpenAI Gym
\citep{Brockman2016-un} API.
Tasks simulated in the environments range from legacy balance-oriented tasks to
simulated robotics and locomotion tasks of various complexities.
In this work, we consider the 5 following environments,
ordered by increasing complexity (degrees of freedom in state and action spaces):
InvertedPendulum,
InvertedDoublePendulum,
Reacher,
Hopper,
Walker2d.
In the experiments presented in \textsc{Figure}~\ref{fig:resplots},
we explore how the performance of SAM
and that of GAIL evolve as a function of the number of interactions
they have with the environment.

For each environment, an expert was designed by training an agent for 10M
timesteps using the Proximal Policy Optimization (PPO) algorithm
\citep{Schulman2017-ou}.
The episode horizon (maximum episode length) was left to its default value per
environment.
We created a dataset of expert trajectories per environment.
For every environment,
we evaluated the performance of the agents
when provided with various quantities of demonstrations,
sampled for the demonstration dataset associated with the environment.
We do so in order to explore how the two methods behave
with respect to the number of demonstrations to which they are exposed.
Both models are shown the same set of selected trajectories.
We ensure that the two
compared models are trained on exactly the same subset of extracted
trajectories by training them with the same random seeds.
We varied the cardinality of the set of selected trajectories
as a function of the environment's complexity.
We ran every experiment on the same range of 4 random seeds, namely
$\{0, 1, 2, 3\}$.
In \textsc{Figure}~\ref{fig:resplots}, we use scatter plots to visualise every
episodic return, for every random seed.
Solid blue and green lines represent the mean episodic return across the
random seeds for the given number of interactions.
The filled areas are confidence intervals around the solid lines,
corresponding to a fixed fraction of the standard deviation around the mean
for the given number of interactions.
Every item coloured in red relates to the expert performance,
for a given environment.
The solid red line corresponds to the mean episodic return of the
demonstrations present in the expert dataset associated with the given
environment.
The filled red region is a trust region whose width is equal to the
standard deviation of returns in the expert dataset.
The dotted line depicts the minimum return in the demonstration dataset while the
dashed line represents the maximum.
Having statistics about the demonstration datasets is particularly insightful
when evaluating the results of experiments dealing with few demonstrations.

Every experiment runs with 4 parallel instantiations of the same model,
initialised with different seeds.
Each instantiation has its own interaction with the environment,
its own replay buffer and its own optimisers.
However, every iteration, the gradients are averaged per module
across instantiations and the
averaged gradients are distributed per module
to every instantiation and immediately used to update the respective module
parameters.
Both \textsc{Sam} and GAIL experiments were run under this setting.
This vertical scalability played a considerable role
in speeding up training phases, equivalently for both models.
Since every instantiation has its own random seed, the fairness of our
performance comparison between \textsc{Sam} and GAIL is further
strengthened \citep{Henderson2017-vm}.

We used layer normalisation \citep{Ba2016-bs} in the policy module.
Indeed, applying layer normalisation to every layer of the policy
was instrumental in yielding better results,
in line with the observations reported in \citep{Plappert2018-rl}.
To ensure symmetry within the actor-critic architecture, we also applied
layer normalisation to the critic module.
\textsc{Pop-Art} \citep{Van_Hasselt2016-bh}
was also useful to our architecture as our learned reward
would sometimes output scores of various magnitudes.
Applying \textsc{Pop-Art} helped in overcoming the various scales.
Finally, note that \textsc{Sam} and GAIL implementations use exactly the same
discriminator implementation.

We provide architecture, hyperparameter, implementation,
and other details in the supplementary material.
We also provide video visualisations of learned policies at
\url{https://youtu.be/-nCsqUJnRKU},
as well as the code associated with this work at
\url{https://github.com/lionelblonde/sam-tf}.

The sample-efficiency we gain over GAIL is considerable:
\textsc{Sam} needs one or two orders of magnitude less interactions with the
environment
to attain asymptotic expert performance.
Note that the horizontal axis is scaled logarithmically.
Additionally, we observe in \textsc{Figure}~\ref{fig:resplots} that
GAIL agents sometimes fall short of reaching the demonstrator's
asymptotic performance (e.g.~Reacher and InverseDoublePendulum).
While GAIL requires full traces of agent--environment interaction
per iteration as it relies on Monte-Carlo estimates,
\textsc{Sam}
only requires a couple of transitions per iteration since it performs
policy evaluation via Temporal Difference learning.
Instead of sampling transitions from the environment, performing an update and
discarding the transitions,
\textsc{Sam} keeps experiential data in memory and can therefore leverage
decorrelated
transitions collected in previous iterations to perform an off-policy update.
Our method therefore requires considerably fewer new samples (interactions)
per iteration, as
it can re-exploit the transitions previously experienced.

Since our approach trades interactions with
the environment with replays with past experiences
to extract more knowledge out of past interactions,
echoing fictitious play in game theory,
it generally takes a longer wall-clock time to train imitation policies.
However, in real-world scenarios (e.g.~robotic
manipulation, autonomous cars), reducing the required interaction with the
world is significantly more desirable, for safety and cost reasons.

\section{Conclusion}

In this work, we introduced a method, called
\textit{Sample-efficient Adversarial Imitation Learning} (\textsc{Sam}),
that meaningfully overcomes one
considerable drawback of
GAIL \citep{Ho2016-bv}:
the number of agent--environment interactions it requires to learn
expert-like policies.
We demonstrate that our method shrinks the number of interactions by an order
of magnitude, and sometimes more.
Leveraging an off-policy procedure was key to that success.

\section*{Acknowledgments}

This work was partially supported by the Swiss National Science
Foundation grant number CRSII5\_177179 ``Modeling pathological gait
resulting from motor impairment'' and European Commission H2020 grant
number \#645220, RAWFIE.


\bibliography{Bib/bibliography}

\end{document}